\documentclass[11pt]{amsproc}
\usepackage{amsthm}
\usepackage{amssymb,latexsym,amsmath}
\usepackage{mathrsfs}
\usepackage{dsfont}
\usepackage{graphicx}
\usepackage{wrapfig}
\usepackage{rotating}
\usepackage{lscape}
\usepackage{float}
\usepackage{tikz}
\usetikzlibrary{arrows}
\theoremstyle{plain}
\newtheorem{T}{Theorem}[section]
\newtheorem{C}{Corollary}[section]

\newtheorem{Pro}{Proposition}[section]
\theoremstyle{remark}
\newtheorem{E}{Example}[section]
\newtheorem{R}{Remark}[section]
\newtheorem{D}{Definition}[section]

\begin{document}

\title{Enhancement of Approximation Spaces by the Use of Primals and Neighborhoods}
\author[A. \c{C}AKSU G\"{U}LER]{A. \c{C}AKSU G\"{U}LER$^1$}%
\keywords{ approximations, primal, rough set } \maketitle

\begin{center}\scriptsize{ Ege University, Faculty of Science, Department of Mathematics, 35100- \.{I}zmir,
Turkey\\
e-mail:aysegul.caksu.guler@ege.edu.tr}

\end{center}

\begin{abstract}
Rough set theory is one of the most widely used and significant approaches for handling incomplete information. It divides the universe in the beginning and uses equivalency relations to produce blocks. Numerous generalized rough set models have been put out and investigated in an effort to increase flexibility and extend the range of possible uses. We introduce four new generalized rough set models that draw inspiration from "neighborhoods and primals" in order to make a contribution to this topic. By minimizing the uncertainty regions, these models are intended to assist decision makers in more effectively analyzing and evaluating the provided data. We verify this goal by demonstrating that the existing models outperform certain current method approaches in terms of improving the approximation operators (upper and lower) and accuracy measurements. We claim that the current models can preserve nearly all significant aspects associated with the rough set model. Preserving the monotonic property, which enables us to assess data uncertainty and boost confidence in outcomes, is one of the intriguing characterizations derived from the existing models. With the aid of specific instances, we also compare the areas of the current approach. Finally, we demonstrate that the new strategy we define for our everyday health-related problem yields more accurate findings.
\end{abstract}

\section{Introduction}
Pawlak \cite{Pawlak1, Pawlak2} introduced rough set theory, which has been thought of as a tool for conceptualizing and analyzing different types of data. It can be applied to attribute value representation models to determine decision rules, assess the importance of attributes, and characterize the connections between them. As a tool for handling ambiguity and uncertainty in information, the idea has significant implications for cognitive sciences and intelligent decision-making. The standard rough set theory was put forward by Pawlak \cite{Pawlak1, Pawlak2}  using two additional sets known as the lower and higher approximations that were derived from an equivalence relationship. The ideas of border regions and accuracy measurements were introduced in order to determine the degree of ambiguity and measure the completeness of the data. The border region can be described as the difference between the approximation operators (upper and lower), whereas the accuracy measure is a number derived from the quotient of the cardinalities of the lower and upper approximations. Many academics substituted similarity relations \cite{dai}, tolerance relations \cite{Pomykala}, or binary relations \cite{Yao2, Zhang} for the equivalence relation in order to get around the severe conditions imposed by the standard rough set model. \

	Yao \cite{Yao1, Yao2, Yao3} initiated this pattern in the 1990s of the previous century in connection to an arbitrary relation. He established new approximation spaces that the left and right neighborhoods induced. The range of applications is increased by these approximation spaces, which also reduce the strict demands of an equivalence relation. However, these models cause some of the original Pawlak model's features to evaporate and, in some circumstances, their accuracy measures exceed one, requiring further treatment \cite{Bably}. Following Yao's breakthrough discoveries, a large number of researchers and academics with an interest in rough set theory offered various kinds of neighborhood systems and used to create new generalized rough set models. In keeping with this, various neighborhood systems have been developed with the intention of introducing a number of generalized rough set models (or approximation spaces), including core neighborhoods \cite{Mareay}, neighborhoods created by the intersection of minimal left and right neighborhoods \cite{Allam1, Allam2}. Instead of utilizing a single binary relation, Abu-Donia\cite{Abu1, Abu2} looked into a family of binary relations to study several approximation operators. Al-shami \cite{Tareq2} recently covered new seven types of rough maximal neighborhoods, which helped to rank COVID-19 suspects and introduce new approximation spaces. \

Kandi et al. \cite{kandil} presented a creative approach to build ASs based on an ideal's structure in 2013. Their goals were to raise accuracy measurements and improve approximation operators. Then, new rough-set models derived from ideal and topological structures were presented by Hosny \cite{Hosny1}. Information systems have been used recently to eliminate uncertainty by using particular neighborhood types with optimal structures\cite{Hosny2, RHosny}.  After that, Güler et al. \cite{caksu} investigated estimates derived from confinement neighborhoods using ideals. Some authors\cite{Tareq2, Hosny3, Hosny4, Hosny5} combined ideals and various maximal neighborhood types to create approximation spaces with desirable qualities to address certain practical problems. \

Acharjee et al. \cite{Acharjee} have presented a novel structure known as a "primal." This structure provides some connections between primal topological spaces and topological spaces in addition to a number of fundamental aspects linked to primals. Filters are the opposite of ideals, while primals appear to be the dual of the grill idea.  But the notions of ideal and primal are unrelated to one another. \

This study intends to present an additional interesting and unique form of approximation spaces derived from ``neighborhoods and primals.'' The main objective in presenting and investigating this version is to raise the subset accuracy levels while improving the approximation operators.
The arrangement of this article is as follows. 
An overview of primals and rough neighborhood systems is provided in Section 2, which is necessary to comprehend this work. Establishing four rough set models and going over their key characterizations is the goal of Section 3. In Section 4, the recently presented models are contrasted with the earlier models to demonstrate their benefits. After that, Section 5 provides a numerical example demonstrating how the present methodology can be successfully used to address a real-world problem.

\section{Preliminaries}
We go through a few essential ideas and results in this section that will be included in this paper.
\begin{D} \cite{kur} 
Let $\Xi$ be a non empty set. Then a family of sets $\mathcal I\subseteq P(\Xi)$  is said to be an ideal in $\Xi$ if
\begin{item}
\item (a)  A, B $\in \mathcal P$ imply $A\cup  B \in$ $\mathcal I$,
\item (b) A $\in \mathcal I$ and B$\subseteq$ A imply B $\in \mathcal I$.
\end{item}
\end{D}
\begin{D} \cite{Acharjee} 
Let $\Xi$ be a non empty set. Then a family of sets $\mathcal P\subseteq P(\Xi)$  is said to be a primal in $\Xi$ if
\begin{item}
\item (a)  $\Xi$ $\not   \in$ $\mathcal P$
\item (b)   $A\cap  B \in$ $\mathcal P$, then A $\in \mathcal P$ or B $\in \mathcal P$
\item (c) A $\in \mathcal P$ and B$\subseteq$ A imply B $ \in \mathcal P$.

\end{item}
\end{D}
The example that follows demonstrates the disconnection between the concepts of ideal and primal.
\begin{E}\label{Example 2.1} 
 Let $\Xi$ = $\{ \iota_1, \iota_2, \iota_3, \iota_4 \}$,  $\mathcal{I}$ = $P(\Xi)$ is ideal but it is not primal. $\mathcal{P}$ = $\{\emptyset,  \{\iota_1 \}, \{\iota_2 \}, \{\iota_4 \}, \{\iota_1, \iota_4 \},  \{\iota_2, \iota_4 \} \}$ is primal but it is not ideal. 
\end{E}
\begin{D} \cite{Yao1, Allam1, Allam2} \label{Definition 2.3}

The following definitions apply to the $w$-neighborhoods of an element $\iota \in \Xi$ that are
 inspired by a relation $\Omega$  : \

$(a)$ after  $\iota$'s neighborhood, shown by $w_a( \iota)$, is obtained as follows: $w_a( \iota)$ = $\{\iota_1  \in \Xi:  (\iota, \iota_1) \in \Omega \}$ \

$(b)$ before  $\iota$'s neighborhood, shown by $w_b( \iota)$, is obtained as follows: $w_b( \iota)$ = $\{\iota_1  \in \Xi:  (\iota_1, \iota) \in \Omega \}$ \

$(c)$ $w_u( \iota)$ = $w_a( \iota)$$\cup$ $w_b( \iota)$ \

$(d)$ $w_i( \iota)$ = $w_a( \iota)$$\cap$ $w_b( \iota)$ \

\end{D}

\begin{D}\cite{Yao2, Yao3} \label{Definition 2.3}
For $w$-neighborhoods and each j $\in$ J =$\{a, b, i, u \}$, the approximation operators (lower and higher), border region, and measurements of accuracy and roughness of a nonempty subset V of $\Xi$ are provided, respectively, by \

  (a) $\underline{N}_{j}(V)= \{ \iota \in \Xi : w_j(\iota) \subseteq V  \}$;\

  (b) $\overline{N}_{j}(V)=  \{ \iota \in \Xi : w_j(\iota) \cap V  \neq \emptyset  \}$.\
  
   (c) $B_{N_{j}}(V)$= $\overline{N}_{j}(V) - \underline{N}_{j}(V) $\

  (d) $\sigma_{N_j}(V)= \frac{|\underline{N}_{j}(V)|}{|\overline{N}_{j}(V)|}$. \\

\end{D}

\section{A FEW NEW ROUGH SET MODELS DERIVED FROM PRIMALS AND $w$-NEIGHBORHOODS}

Four different kinds of rough set models that are defined by primals and neighbors under any arbitrary relation are shown in this section. To shed light on the discovered facts and relationships, various counterexamples are included in addition to a critical examination of their primary attributes and characterizations.
\subsection{PRIMAL-BASED GENERALIZED ROUGH SETS: THE FIRST TECHNIQUE}
\begin{D}\label{Definition 3.1} 
Let  $\mathcal{P}$  and $\Omega$  stand for the primal on a nonempty set $\Xi$ and the binary relation, respectively. The boundary region, accuracy, roughness, and enhanced operators (upper and lower) of a nonempty subset V of $\Xi$ generated from $\Omega$ and $\mathcal{P}$  each j $\in$ J  are provided, respectively, by \

  (a) $\underline{N1}_{j}^{\mathcal{P}}(V)= \{ \iota \in \Xi : w_j(\iota) \cap V ' \in \mathcal{P} \}$;\

  (b) $\overline{N1}_{j}^{\mathcal{P}}(V)=  \{ \iota \in \Xi : w_j(\iota) \cap V  \not \in \mathcal{P}  \}$.\
  
   (c) $B^{\mathcal{P}}_{N1_{j}}(V)$= $\overline{N1}^{\mathcal{P}}_{j}(V) - \underline{N1}^{\mathcal{P}}_{j}(V) $\

  (d) $\sigma_{N1_j}^\mathcal{P}(V)= \frac{|\underline{N1}^{\mathcal{P}}_{j}(V)) \cap V|}{|\overline{N1}^{\mathcal{P}}_{j}(V)) \cup V|}$. \\

 If $\mathcal{P}$ =$\{\emptyset \}$ then $\underline{N1}_{j}^{\mathcal{P}}(V)$ = $\underline{N}_{j}(V)$, $\overline{N1}_{j}^{\mathcal{P}}(V)$ = $\overline{N}_{j}(V)$ and $\sigma_{N1_j}^\mathcal{P}(V)$= $\sigma_{N1_j}(V)$.
 
\end{D}

\begin{Pro} \label{Proposition 3.1} 
Let $\mathcal{P}$, $\mathcal{P}1$ be primals and $\Omega$ be a binary relation on $\Xi$. Then, let V, W $\subseteq$  $\Xi$. Then, the subsequent is true :\

  $ (a)$   $\underline{N1}_{j}^{\mathcal{P}}(\Xi)$ = $\Xi$ and $\overline{N1}_{j}^{\mathcal{P}}(\emptyset)$ = $\emptyset$

  $ (b)$ $V\subseteq W$ implies $\underline{N1}_{j}^{\mathcal{P}}(V)$$\subseteq$ $\underline{N1}_{j}^{\mathcal{P}}(W)$;\

    \indent \indent \indent \indent  \indent \indent $\overline{N1}_{j}^{\mathcal{P}}(V)$ $\subseteq $$\overline{N1}_{j}^{\mathcal{P}}(W)$;\

  $ (c)$ $\underline{N1}_{j}^{\mathcal{P}}(V \cup W)\supseteq$ $\underline{N1}_{j}^{\mathcal{P}}(V) \cup$ $\underline{N1}_{j}^{\mathcal{P}}(W)$;\

   \indent \indent   $\overline{N1}_{j}^{\mathcal{P}}(V \cup W)\supseteq$ $\overline{N1}_{j}^{\mathcal{P}}(V) \cup$ $\overline{N1}_{j}^{\mathcal{P}}(W)$;\

  $ (d)$ $\underline{N1}_{j}^{\mathcal{P}}(V \cap W)\subseteq$ $\underline{N1}_{j}^{\mathcal{P}}(V) \cap$ $\underline{N1}_{j}^{\mathcal{P}}(W)$;\

   \indent \indent   $\overline{N1}_{j}^{\mathcal{P}}(V \cap W)\subseteq$ $\overline{N1}_{j}^{\mathcal{P}}(V) \cap$ $\overline{N1}_{j}^{\mathcal{P}}(W)$;\

$(e)$ $\underline{N1}_{j}^{\mathcal{P}}(V)$ = $(\overline{N1}_{j}^{\mathcal{P}}(V '))'$

   \begin{proof} \

$(a)$  The proof by Definition 3.1 is simple.\

 $ (b)$ Let  $ \iota \in \underline{N1}_{j}^{\mathcal{P}}(V)$. Then $w_j(\iota) \cap V ' \in \mathcal{P}$. Since  $V \subseteq W$ and  by the definition of primal, $ w_j(\iota) \cap W' \in \mathcal{P}$. So 
$\underline{N1}_{j}^{\mathcal{P}}(V)$ $\subseteq$ 
$\underline{N1}_{j}^{\mathcal{P}}(W)$. \

 $ (c)$ It originates directly from  $(b)$. \

 $ (d)$ It originates directly from  $(b)$. \

$(e)$  $(\overline{N1}_{j}^{\mathcal{P}}(V '))'$ = $\{ \iota \in \Xi : w_j(\iota) \cap V'  \not \in \mathcal{P}  \}'$ = $\{ \iota \in \Xi : w_j(\iota) \cap V'  \in \mathcal{P}  \}$ =  $\underline{N1}_{j}^{\mathcal{P}}(V )$

\end{proof}
\end{Pro}

\begin{Pro} \label{Proposition 3.2} 
Let $\mathcal{P}$, $\mathcal{P}1$ be primals and $\Omega$ be a binary relation on $\Xi$. Then, let V $\subseteq$  $\Xi$. Then, the subsequent is true :\

$ (a)$  If $\mathcal{P}$ = $P(\Xi) - \{ \Xi \}$ then 

if  V $\neq$ $\Xi$ then $\underline{N1}_{j}^{\mathcal{P}}(V)$ = $\emptyset$ and 
if V $\subseteq$ $\Xi$ then $\overline{N1}_{j}^{\mathcal{P}}(V)$ = $\emptyset$  \

   $(b)$  If $ V^{ı}$ $\in$ $\mathcal{P}$  then $\underline{N1}_{j}^{\mathcal{P}}(V)$ = $\Xi$ \

  $(c)$ If V $\in$ $\mathcal{P}$  then $\overline{N1}_{j}^{\mathcal{P}}(V)$ =  $\emptyset$; \

$(d)$ If  $\mathcal{P}$ $\subseteq$ $\mathcal{P}1$ then 
$\underline{N1}_{j}^{\mathcal{P}}(V)$$\subseteq$ $\underline{N1}_{j}^{\mathcal{P}1}(V)$; \

\indent \indent  \indent \indent \indent  \indent
$\overline{N1}_{j}^{\mathcal{P}1}(V)$$\subseteq$ $\overline{N1}_{j}^{\mathcal{P}}(V)$ \

   $(e)$ $\overline{N1}_{j}^{\mathcal{P}\cup \mathcal{P}1}(V )$ = $\overline{N1}_{j}^{\mathcal{P}}(V) \cap$ $\overline{N1}_{j}^{\mathcal{P}1}(V)$.\

   \begin{proof} \

$(a)$  Since V $\neq$ $\Xi$ and $\mathcal{P}$ = $P(\Xi) - \{ \Xi \}$,  we obtain that $\overline{N1}_{j}^{\mathcal{P}}(V)=  \{ \iota \in \Xi : w_j(\iota) \cap V  \not \in \mathcal{P}  \}$ = $\emptyset$.\

$(b)$  It originates directly from  definition of primal.

$(e)$ $\overline{N1}_{j}^{\mathcal{P}\cup \mathcal{P}1}(V ) $= $\{ \iota \in \Xi : w_j(\iota) \cap V  \not \in \mathcal{P}\cup \mathcal{P}1  \}$ = $\{ \iota \in \Xi : w_j(\iota) \cap V  \not \in \mathcal{P}  \}$ $\cap$ $\{ \iota \in \Xi : w_j(\iota) \cap V  \not \in  \mathcal{P}1  \}$ = $\overline{N1}_{j}^{\mathcal{P}}(V ) $ $\cap$ $\overline{N1}_{j}^{\mathcal{P}1}(V ) $. \

\end{proof}
\end{Pro}

The equations in Proposition \ref{Proposition 3.1}(b), (c) and (d) might not hold true in general, as demonstrated by the example that follows. Furthermore, it suggests that Proposition \ref{Proposition 3.2} (b) and (c) might not deliver their contradictory meanings.

\begin{E}\label{Example 3.1} 
 Let $\Xi$ = $\{ \iota_1, \iota_2, \iota_3, \iota_4 \}$,  $\Omega$ = $\{ (\iota_1, \iota_1), (\iota_2, \iota_2), (\iota_3, \iota_3), (\iota_1, \iota_2), (\iota_2, \iota_3) \}$ and $\mathcal{P}$ = $\{\emptyset,  \{\iota_1 \}, \{\iota_2 \}, \{\iota_3 \}, \{\iota_1, \iota_3 \},  \{\iota_2, \iota_3 \} \}$. \

$(a)$ Let $V$ = $\{  \iota_1, \iota_2, \iota_3 \}$ and $W$ = $\{  \iota_3, \iota_4 \}$. Then $\underline{N1}_{a}^{\mathcal{P}}(V)$ = $\Xi$ and $\underline{N1}_{a}^{\mathcal{P}}(W)$ = $\{ \iota_2, \iota_3, \iota_4 \}$. We gain $\underline{N1}_{a}^{\mathcal{P}}(W)$$\subseteq$ $\underline{N1}_{a}^{\mathcal{P}}(V)$ but it is not  $W \not \subseteq V$. \

$(b)$ Let $V$ = $\{  \iota_3, \iota_4 \}$. Then $\overline{N1}_{a}^{\mathcal{P}}(V)$ = $\emptyset$ but it is not  $V \not \in \mathcal{P}$. \

\end{E}
\begin{E}\label{Example 3.2} 
 Let $\Xi$ = $\{ \iota_1, \iota_2, \iota_3, \iota_4 \}$,  $\Omega$ = $\{ (\iota_1, \iota_1), (\iota_2, \iota_2), (\iota_3, \iota_3), (\iota_1, \iota_2),  (\iota_1, \iota_3), \\ (\iota_2, \iota_1),  (\iota_3, \iota_1),  (\iota_4, \iota_2) \}$ and $\mathcal{P}$ = $\{\emptyset,  \{\iota_1 \}, \{\iota_4 \}\}$. \

$(a)$ Let $V$ = $\{  \iota_1, \iota_2  \}$ and $W$ = $\{  \iota_1, \iota_3 \}$. Then 
$\overline{N1}_{a}^{\mathcal{P}}(V)$ = $\{ \iota_1, \iota_2, \iota_4 \}$ and $\overline{N1}_{j}^{\mathcal{P}}(W)$ = $\{ \iota_1, \iota_3 \}$. We gain $\overline{N1}_{a}^{\mathcal{P}}(V)$$\cap$ $\overline{N1}_{a}^{\mathcal{P}}(W)$$\neq$ $\overline{N1}_{a}^{\mathcal{P}}(V \cap W)$. \

$(b)$ Let $V$ = $\{  \iota_1, \iota_3  \}$ and $W$ = $\{  \iota_2, \iota_4 \}$. Then 
$\underline{N1}_{a}^{\mathcal{P}}(V)$ = $\{  \iota_3  \}$ and $\underline{N1}_{j}^{\mathcal{P}}(W)$ = $\{  \iota_2, \iota_3 , \iota_4 \}$. We gain $\underline{N1}_{a}^{\mathcal{P}}(V)$$\cup$ $\underline{N1}_{a}^{\mathcal{P}}(W)$$\neq$ $\underline{N1}_{a}^{\mathcal{P}}(V \cup W)$. \
  
$(c)$ Let $V$ = $\{  \iota_1, \iota_4  \}$. Then 
$\underline{N1}_{a}^{\mathcal{P}}(V)$ = $\emptyset$ but $V^{ı}$ $ \not \in$ $\mathcal{P}$.

\end{E}
\begin{R} \label{Remark 3.2} 
  The current model lacks the following characteristics of the approximation operators in the Pawlak model. \

$(a)$ $\underline{N1}_{j}^{\mathcal{P}}(V) \subseteq V \subseteq \overline{N1}_{j}^{\mathcal{P}}(V)$; \

$(b)$  $\underline{N1}_{j}^{\mathcal{P}}(\underline{N1}_{j}^{\mathcal{P}}(V))= \underline{N1}_{j}^{\mathcal{P}}(V);$ \

 \indent \indent  $\overline{N1}_{j}^{\mathcal{P}}(\overline{N1}_{j}^{\mathcal{P}}(V))= \overline{N1}_{j}^{\mathcal{P}}(V);$  \

 $(c)$  $\underline{N1}_{j}^{\mathcal{P}}(V \cap W) =$ $\underline{N1}_{j}^{\mathcal{P}}(V) $
$\cap$ $\underline{N1}_{j}^{\mathcal{P}}(W)$;\

$ (d)$  $\overline{N1}_{j}^{\mathcal{P}}(V \cup W)=$ $\overline{N1}_{j}^{\mathcal{P}}(V)$
$ \cup$ $\overline{N1}_{j}^{\mathcal{P}}(W)$;\ 

$(e)$ $\overline{N1}_{j}^{\mathcal{P}}(\underline{N1}_{j}^{\mathcal{P}}(V))= \underline{N1}_{j}^{\mathcal{P}}(V)$;\

 \indent \indent  $\underline{N1}_{j}^{\mathcal{P}}(\overline{N1}_{j}^{\mathcal{P}}(V))= \overline{N1}_{j}^{\mathcal{P}}(V)$
\end{R}
\begin{E}\label{Example 3.3} 
 Let $\Xi$ = $\{ \iota_1, \iota_2, \iota_3, \iota_4 \}$,  $\Omega$ = $\{ (\iota_1, \iota_1), (\iota_2, \iota_3), (\iota_2, \iota_4),  (\iota_3, \iota_1), (\iota_3, \iota_4) \}$ and $\mathcal{P}$ = $\{\emptyset,  \{\iota_1 \}, \{\iota_4 \}\}$. \

$(a)$ Let $V$ = $\{  \iota_1, \iota_3  \}$. Then  $\underline{N1}_{j}^{\mathcal{P}}(V)$ = $\Xi$.  So  $\underline{N1}_{a}^{\mathcal{P}}(V) \not \subseteq V$; 

$(b)$ Let $V$ = $\{  \iota_2,  \iota_3, \iota_4 \}$ and $W$ = $\{  \iota_1, \iota_2 , \iota_3\}$. Then $\underline{N1}_{a}^{\mathcal{P}}(V)$ = $\Xi$, $\underline{N1}_{a}^{\mathcal{P}}(W)$ = $\Xi$ and  $\underline{N1}_{a}^{\mathcal{P}}(V \cap W)$ =  $\{  \iota_1, \iota_2, \iota_4\}$ .  So  $\underline{N1}_{a}^{\mathcal{P}}(V \cap W) \neq$ $\underline{N1}_{a}^{\mathcal{P}}(V) $
$\cap$ $\underline{N1}_{a}^{\mathcal{P}}(W)$.\
\end{E}
\begin{E}\label{Example 3.4} 
 Let $\Xi$ = $\{ \iota_1, \iota_2, \iota_3, \iota_4 \}$,  $\Omega$ = $\{ (\iota_1, \iota_1), (\iota_2, \iota_2), (\iota_3, \iota_3), (\iota_4, \iota_4), (\iota_1, \iota_2),  (\iota_1, \iota_3), (\iota_2, \iota_3), \\  (\iota_3, \iota_4),  (\iota_4, \iota_1) \}$ and $\mathcal{P}$ = $\{\emptyset,  \{\iota_2 \}, \{\iota_3 \}\}$. \

$(a)$ Let $V$ = $\{  \iota_1, \iota_3  \}$. Then 
$\overline{N1}_{a}^{\mathcal{P}}(V)$ = $\{  \iota_1, \iota_4  \}$ and 
$\overline{N1}_{a}^{\mathcal{P}}(\overline{N1}_{a}^{\mathcal{P}}(V))$ =$\{ \iota_1, \iota_3, \iota_4 \}$. Then 
$\overline{N1}_{a}^{\mathcal{P}}(\overline{N1}_{a}^{\mathcal{P}}(V)) \neq \overline{N1}_{a}^{\mathcal{P}}(V)$. \

$(b)$ Let $V$ = $\{  \iota_1, \iota_3  \}$. Then 
$\underline{N1}_{a}^{\mathcal{P}}(V)$ = $\{ \iota_1, \iota_2 \}$ and 
$\overline{N1}_{a}^{\mathcal{P}}(\underline{N1}_{a}^{\mathcal{P}}(V))$ =$\{ \iota_1, \iota_4 \}$. Then 
$\overline{N1}_{a}^{\mathcal{P}}(\underline{N1}_{a}^{\mathcal{P}}(V))\neq \underline{N1}_{a}^{\mathcal{P}}(V)$. \
\end{E}

\begin{Pro} \label{Proposition 3.2} 
 The followings are hold:\

$(a)$ $\underline{N1}_{u}^{\mathcal{P}}(V)$$\subseteq$ $\underline{N1}_{a}^{\mathcal{P}}(V)$
$\subseteq$ $\underline{N1}_{i}^{\mathcal{P}}(V)$;\\

 \indent \indent $\overline{N1}_{i}^{\mathcal{P}}(V)$$\subseteq$ $\overline{N1}_{a}^{\mathcal{P}}(V)$
$\subseteq$ $\overline{N1}_{u}^{\mathcal{P}}(V)$;\\

$(b)$ $\underline{N1}_{u}^{\mathcal{P}}(V)$$\subseteq$ $\underline{N1}_{b}^{\mathcal{P}}(V)$$\subseteq$ $\underline{N1}_{i}^{\mathcal{P}}(V)$;\\

 \indent \indent $\overline{N1}_{i}^{\mathcal{P}}(V)$$\subseteq$ $\overline{N1}_{b}^{\mathcal{P}}(V)$
$\subseteq$ $\overline{N1}_{u}^{\mathcal{P}}(V)$;\\

\begin{proof} \
It is obvious.
\end{proof}
\end{Pro}

\begin{C} \label{Corollary 3.1} 
 The followings are hold:\

$ (a)$ $B^{\mathcal{P}}_{N1_{i}}(V)$$\subseteq$$B^{\mathcal{P}}_{N1_{a}}(V)$$\subseteq$$B^{\mathcal{P}}_{N1_{u}}(V)$;\\

$ (b)$ $B^{\mathcal{P}}_{N1_{i}}(V)$$\subseteq$$B^{\mathcal{P}}_{N1_{b}}(V)$$\subseteq$$B^{\mathcal{P}}_{N1_{u}}(V)$.\\

\end{C}

\begin{C} \label{Corollary 3.2} 
Let $\mathcal{P}$ be primal and $\Omega$ be a binary relation on $\Xi$. Then, let V
$\subseteq$  $\Xi$. Then, the subsequent is true :\\

(a) $\sigma_{N1_u}^\mathcal{P}(V) $$\leq$ $\sigma_{N1_a}^\mathcal{P}(V)$$\leq$ 
$\sigma_{N1_i}^\mathcal{P}(V)$ ;\\

(b) $\sigma_{N1_u}^\mathcal{P}(V) $$\leq$ $\sigma_{N1_b}^\mathcal{P}(V)$$\leq$ 
$\sigma_{N1_i}^\mathcal{P}(V)$.\\

\end{C}

\begin{C} \label{Corollary 3.3} 
    Let $\mathcal{P}$, $\mathcal{P}1$ be primals such that $\mathcal{P}$ $\supseteq$ $\mathcal{P}1$  and $\Omega$ be a binary relation on $\Xi$. Then, let V $\subseteq$  $\Xi$. Then, the subsequent is true :\\

   $ (a)$ $B^{\mathcal{P}1}_{N1_{j}}(V)$$\subseteq$$B^{\mathcal{P}}_{N1_{j}}(V)$\

   (b) $\sigma_{N1_j}^{\mathcal{P}1}(V) $$\leq$ $\sigma_{N1_j}^\mathcal{P}(V)$.
\end{C}

\begin{E}\label{Example 3.5} 
Consider Example 3.1. $\mathcal{P}1$ = $\{\emptyset,  \{\iota_1 \}, \{\iota_2 \} \}$. 
 Let $V$ = $\{  \iota_3, \iota_4 \}$. Then $\sigma_{N1_a}^ \mathcal{P}(V)$ = 1 but 
$\sigma_{N1_a}^{\mathcal{P}1}(V)$ =$\frac{2}{3}$.
\end{E}
\subsection{PRIMAL-BASED GENERALIZED ROUGH SETS: THE SECOND TECHNIQUE}
\begin{D}\label{Definition 3.2} 
Let $\mathcal{P}$  and $\Omega$ stand for the primal on a nonempty set $\Xi$ and the binary relation, respectively. The border region, accuracy, roughness, and enhanced operators (upper and lower) of a nonempty subset V of$\Xi$ generated from $\Omega$ and $\mathcal{P}$ are provided, respectively, by \

  (a) $\underline{N2}_{j}^{\mathcal{P}}(V)= \{ \iota \in \Xi : w_j(\iota) \cap V ' \in \mathcal{P} \}$;\

  (b) $\overline{N2}_{j}^{\mathcal{P}}(V)$=  V $\cup$ $\overline{N1}_{j}^{\mathcal{P}}(V)$
  
   (c) $B^{\mathcal{P}}_{N2_{j}}(V)$= $\overline{N2}^{\mathcal{P}}_{j}(V) - \underline{N2}^{\mathcal{P}}_{j}(V) $\

  (d) $\sigma_{N2_j}^\mathcal{P}(V)= \frac{|\underline{N2}^{\mathcal{P}}_{j}(V)\cap V |}{|\overline{N2}^{\mathcal{P}}_{j}(V) |}$. \\

\end{D}

\begin{Pro} \label{Proposition 3.4} 
Let$\mathcal{P}$, $\mathcal{P}1$ be primals and $\Omega$be a binary relation on $\Xi$. Then, let V, W $\subseteq$  $\Xi$. Then, the subsequent is true :\

  $ (a)$  V  $\subseteq$$\overline{N2}_{j}^{\mathcal{P}}(V)$;

  $ (b)$   $\underline{N2}_{j}^{\mathcal{P}}(\Xi)$ = $\Xi$ and $\overline{N2}_{j}^{\mathcal{P}}(\emptyset)$ = $\emptyset$;

  $ (c)$ $V\subseteq W$ implies $\underline{N2}_{j}^{\mathcal{P}}(V)$$\subseteq$ $\underline{N2}_{j}^{\mathcal{P}}(W)$;\

    \indent \indent \indent \indent  \indent \indent $\overline{N2}_{j}^{\mathcal{P}}(V)$ $\subseteq $$\overline{N2}_{j}^{\mathcal{P}}(W)$;\

  $ (c)$ $\underline{N2}_{j}^{\mathcal{P}}(V \cup W)\supseteq$ $\underline{N2}_{j}^{\mathcal{P}}(V) \cup$ $\underline{N2}_{j}^{\mathcal{P}}(W)$;\

   \indent \indent   $\overline{N2}_{j}^{\mathcal{P}}(V \cup W)\supseteq$ $\overline{N2}_{j}^{\mathcal{P}}(V) \cup$ $\overline{N2}_{j}^{\mathcal{P}}(W)$;\

  $ (d)$ $\underline{N2}_{j}^{\mathcal{P}}(V \cap W)\subseteq$ $\underline{N2}_{j}^{\mathcal{P}}(V) \cap$ $\underline{N2}_{j}^{\mathcal{P}}(W)$;\

   \indent \indent   $\overline{N2}_{j}^{\mathcal{P}}(V \cap W)\subseteq$ $\overline{N2}_{j}^{\mathcal{P}}(V) \cap$ $\overline{N2}_{j}^{\mathcal{P}}(W)$;\

   \begin{proof} \ Similiar to Proposition 3.1.

\end{proof}
\end{Pro}

\begin{Pro} \label{Proposition 3.5} 
Let$\mathcal{P}$, $\mathcal{P}1$ be primals and $\Omega$be a binaryrelation on $\Xi$. Then, let V$\subseteq$  $\Xi$. Then, the subsequent is true :\

$ (a)$  If $\mathcal{P}$ = $P(\Xi) - \{ \Xi \}$ then 

if  V $\neq$ $\Xi$ then $\underline{N2}_{j}^{\mathcal{P}}(V)$ = $\emptyset$ and 
if V $\subseteq$ $\Xi$ then $\overline{N2}_{j}^{\mathcal{P}}(V)$ = V.  \

   $(b)$  If $V^{ı}$ $\in$ $\mathcal{P}$  then $\underline{N2}_{j}^{\mathcal{P}}(V)$ = $\Xi$ \

  $(c)$ If V $\in$ $\mathcal{P}$  then $\overline{N2}_{j}^{\mathcal{P}}(V)$ =  V; \

$(d)$ If  $\mathcal{P}$ $\subseteq$ $\mathcal{P}1$ then 
$\underline{N2}_{j}^{\mathcal{P}}(V)$$\subseteq$ $\underline{N2}_{j}^{\mathcal{P}1}(V)$; \

\indent \indent  \indent \indent \indent  \indent
$\overline{N2}_{j}^{\mathcal{P}1}(V)$$\subseteq$ $\overline{N2}_{j}^{\mathcal{P}}(V)$ \

   $(e)$ $\overline{N2}_{j}^{\mathcal{P}\cup \mathcal{P}1}(V )$ = $\overline{N2}_{j}^{\mathcal{P}}(V) \cap$ $\overline{N2}_{j}^{\mathcal{P}1}(V)$;\

   \begin{proof} \

 Similiar to Proposition 3.2.

\end{proof}
\end{Pro}

The equations in Proposition \ref{Proposition 3.4}(b), (c) and (d) might not hold true in general, as demonstrated by the example that follows. Furthermore, it suggests that Proposition \ref{Proposition 3.5} (b) and (c) might not deliver their contradictory meanings.

\begin{E}\label{Example 3.6} 
 Let $\Xi$ = $\{ \iota_1, \iota_2, \iota_3, \iota_4 \}$,  $\Omega$ = $\{ (\iota_1, \iota_1), (\iota_2, \iota_3),  (\iota_2, \iota_4),  (\iota_3, \iota_2)\}$ and $\mathcal{P}$ = $\{\emptyset,  \{\iota_1 \}, \{\iota_4 \}\}$. \

$(a)$ Let $V$ = $\{  \iota_1, \iota_2  \}$ and $W$ = $\{  \iota_1, \iota_3, \iota_4 \}$. Then 
$\overline{N2}_{a}^{\mathcal{P}}(V)$ = $\{ \iota_1, \iota_2, \iota_3 \}$ and $\overline{N2}_{a}^{\mathcal{P}}(W)$ = $\Xi$. We gain $\overline{N2}_{a}^{\mathcal{P}}(V)$$\subseteq$ $\overline{N2}_{a}^{\mathcal{P}}(W)$ but it is not  $V \not \subseteq W$. \

$(b)$ Let $V$ = $\{  \iota_1, \iota_4 \}$. Then $\overline{N2}_{a}^{\mathcal{P}}(V)$ = $\{  \iota_1, \iota_4 \}$ but it is not  $V \not \in \mathcal{P}$. \

$(c)$ Let $V$ = $\{  \iota_1, \iota_2  \}$ and $W$ = $\{  \iota_1, \iota_3, \iota_4 \}$. Then 
$\overline{N2}_{a}^{\mathcal{P}}(V)$ = $\{ \iota_1, \iota_2, \iota_3 \}$ and $\overline{N2}_{j}^{\mathcal{P}}(W)$ = $\Xi$.  We gain  $\overline{N2}_{a}^{\mathcal{P}}(V)$$\cap$ $\overline{N2}_{a}^{\mathcal{P}}(W)$$\neq$ $\overline{N2}_{a}^{\mathcal{P}}(V \cap W)$. \

\end{E}

\begin{R} \label{Remark 3.2} 
 The current model lacks the following characteristics of the approximation operators in the Pawlak model. \

$(a)$ $\underline{N2}_{j}^{\mathcal{P}}(V) \subseteq V $; \

$(b)$  $\underline{N2}_{j}^{\mathcal{P}}(\underline{N2}_{j}^{\mathcal{P}}(V))= \underline{N2}_{j}^{\mathcal{P}}(V);$ \

 \indent \indent  $\overline{N2}_{j}^{\mathcal{P}}(\overline{N2}_{j}^{\mathcal{P}}(V))= \overline{N2}_{j}^{\mathcal{P}}(V);$  \

 $(c)$  $\underline{N2}_{j}^{\mathcal{P}}(V \cap W) =$ $\underline{N2}_{j}^{\mathcal{P}}(V) $
$\cap$ $\underline{N2}_{j}^{\mathcal{P}}(W)$;\

$ (d)$  $\overline{N2}_{j}^{\mathcal{P}}(V \cup W)=$ $\overline{N2}_{j}^{\mathcal{P}}(V)$
$ \cup$ $\overline{N2}_{j}^{\mathcal{P}}(W)$;\ 

$(e)$ $\overline{N2}_{j}^{\mathcal{P}}(\underline{N2}_{j}^{\mathcal{P}}(V))= \underline{N2}_{j}^{\mathcal{P}}(V)$;\

 \indent \indent  $\underline{N2}_{j}^{\mathcal{P}}(\overline{N2}_{j}^{\mathcal{P}}(V))= \overline{N2}_{j}^{\mathcal{P}}(V)$

$(f)$ $\underline{N2}_{j}^{\mathcal{P}}(V)$ = $(\overline{N2}_{j}^{\mathcal{P}}(V '))'$
\end{R}
\begin{E}\label{Example 3.7} 
 Let $\Xi$ = $\{ \iota_1, \iota_2, \iota_3, \iota_4 \}$,  $\Omega$ = $\{ (\iota_1, \iota_1), (\iota_2, \iota_3),  (\iota_2, \iota_4),  (\iota_3, \iota_2)\}$ and $\mathcal{P}$ = $\{\emptyset,  \{\iota_1 \}, \{\iota_4 \}\}$. \

$(a)$ Let $V$ = $\{  \iota_1, \iota_2  \}$. Then  $\underline{N2}_{a}^{\mathcal{P}}(V)$ = 
$\{  \iota_1, \iota_3 , \iota_4 \}$.  $\overline{N2}_{a}^{\mathcal{P}}(V ')$ = $\{  \iota_2, \iota_3 , \iota_4 \}$  $\underline{N2}_{a}^{\mathcal{P}}(V)$ $\neq$ $(\overline{N2}_{a}^{\mathcal{P}}(V '))'$; \

$(b)$ Let $V$ = $\{  \iota_1, \iota_2  \}$. Then 
$\underline{N2}_{a}^{\mathcal{P}}(V)$ = $\{ \iota_1, \iota_3, \iota_4 \}$ and 
$\overline{N2}_{a}^{\mathcal{P}}(\underline{N2}_{a}^{\mathcal{P}}(V))$ =$\Xi$. Then 
$\overline{N2}_{a}^{\mathcal{P}}(\underline{N2}_{a}^{\mathcal{P}}(V))\neq \underline{N2}_{a}^{\mathcal{P}}(V)$.

\end{E}
\begin{E}\label{Example 3.8} 
Examine Example 3.4. Let $V$ = $\{  \iota_1  \}$ and  $W$ = $\{  \iota_2, \iota_3 \}$. Then 
$\overline{N2}_{a}^{\mathcal{P}}(V)$ =$\{  \iota_1, \iota_3  \}$, 
$\overline{N2}_{a}^{\mathcal{P}}(W)$ =$\{  \iota_1, \iota_2, \iota_3  \}$ and 
$\overline{N2}_{a}^{\mathcal{P}}(V \cup W)$=$\Xi$. So $\overline{N2}_{a}^{\mathcal{P}}(V \cup W)$$\neq$ $\overline{N2}_{a}^{\mathcal{P}}(V)$
$ \cup$ $\overline{N2}_{a}^{\mathcal{P}}(W)$.
\end{E}
\begin{E}\label{Example 3.9} 
 Let $\Xi$ = $\{ \iota_1, \iota_2, \iota_3, \iota_4 \}$,  $\Omega$ = $\{ (\iota_1, \iota_1), (\iota_2, \iota_1),  (\iota_2, \iota_3),  (\iota_3, \iota_2), (\iota_4, \iota_2)\}$ and $\mathcal{P}$ = $\{\emptyset,  \{\iota_1 \}, \{\iota_4 \}\}$.  Let $V$ = $\{  \iota_3 \}$. Then  $\overline{N2}_{b}^{\mathcal{P}}(V)$ = $\{  \iota_2, \iota_3 \}$. $\overline{N2}_{b}^{\mathcal{P}}(\overline{N2}_{b}^{\mathcal{P}}(V))$ = $\{  \iota_1, \iota_2 , \iota_3 \}$.  So  $\overline{N2}_{b}^{\mathcal{P}}(\overline{N2}_{b}^{\mathcal{P}}(V)) $$\neq$ 
$\overline{N2}_{b}^{\mathcal{P}}(V)$.
\end{E}
\begin{Pro} \label{Proposition 3.6} 
 The followings are hold:\

$(a)$ $\underline{N2}_{u}^{\mathcal{P}}(V)$$\subseteq$ $\underline{N2}_{a}^{\mathcal{P}}(V)$
$\subseteq$ $\underline{N2}_{i}^{\mathcal{P}}(V)$;\\

 \indent \indent $\overline{N2}_{i}^{\mathcal{P}}(V)$$\subseteq$ $\overline{N2}_{a}^{\mathcal{P}}(V)$
$\subseteq$ $\overline{N2}_{u}^{\mathcal{P}}(V)$;\\

$(b)$ $\underline{N2}_{u}^{\mathcal{P}}(V)$$\subseteq$ $\underline{N2}_{b}^{\mathcal{P}}(V)$$\subseteq$ $\underline{N2}_{i}^{\mathcal{P}}(V)$;\\

 \indent \indent $\overline{N2}_{i}^{\mathcal{P}}(V)$$\subseteq$ $\overline{N2}_{b}^{\mathcal{P}}(V)$
$\subseteq$ $\overline{N2}_{u}^{\mathcal{P}}(V)$;\\

\begin{proof} \
It is obvious.
\end{proof}
\end{Pro}

\begin{C} \label{Corollary 3.1} 
The subsequent items are hold:\

$ (a)$ $B^{\mathcal{P}}_{N2_{i}}(V)$$\subseteq$$B^{\mathcal{P}}_{N2_{a}}(V)$$\subseteq$$B^{\mathcal{P}}_{N2_{u}}(V)$;\\

$ (b)$ $B^{\mathcal{P}}_{N2_{i}}(V)$$\subseteq$$B^{\mathcal{P}}_{N2_{b}}(V)$$\subseteq$$B^{\mathcal{P}}_{N2_{u}}(V)$.\\

\end{C}

\begin{C} \label{Corollary 3.2} 
Let$\mathcal{P}$ be primal and $\Omega$be a binary relation on $\Xi$. Then, let V
$\subseteq$  $\Xi$. Then, the subsequent is true :\\

(a) $\sigma_{N2_u}^\mathcal{P}(V) $$\leq$ $\sigma_{N2_a}^\mathcal{P}(V)$$\leq$ 
$\sigma_{N2_i}^\mathcal{P}(V)$ ;\\

(b) $\sigma_{N2_u}^\mathcal{P}(V) $$\leq$ $\sigma_{N2_b}^\mathcal{P}(V)$$\leq$ 
$\sigma_{N2_i}^\mathcal{P}(V)$.\\

\end{C}

\begin{C} \label{Corollary 3.3} 
    Let$\mathcal{P}$, $\mathcal{P}1$ be primals such that $\mathcal{P}$ $\supseteq$$\mathcal{P}1$  and $\Omega$be a binaryrelation on $\Xi$. Then, let V $\subseteq$  $\Xi$. Then, the subsequent is true :\\

   $ (a)$ $B^{\mathcal{P}1}_{N2_{j}}(V)$$\subseteq$$B^{\mathcal{P}}_{N2_{j}}(V)$\

   (b) $\sigma_{N2_j}^{\mathcal{P}1}(V) $$\leq$ $\sigma_{N2_j}^\mathcal{P}(V)$.
\end{C}

\begin{E}\label{Example 3.1} 
Consider Example 3.9. $\mathcal{P}1$ = $\{\emptyset,  \{\iota_1 \} \}$. 
 Let $V$ = $\{ \iota_4 \}$. Then $\sigma_{N2_b}^ \mathcal{P}(V)$ = 1 but 
$\sigma_{N2_b}^{\mathcal{P}1}(V)$ =$\frac{1}{2}$.
\end{E}

\subsection{PRIMAL-BASED GENERALIZED ROUGH SETS: THE THIRD TECHNIQUE}
\begin{D}\label{Definition 3.3} 
Let $\mathcal{P}$  and $\Omega$ stand for the primal on a nonempty set $\Xi$ and the binary relation, respectively. The border region, accuracy, roughness, and enhanced operators (upper and lower) of a nonempty subset V of$\Xi$ generated from $\Omega$ and $\mathcal{P}$ are provided, respectively, by \

  (a) $\underline{N3}_{j}^{\mathcal{P}}(V)=\bigcup_{\iota \in \Xi} \{ w_j(\iota) : w_j(\iota) \cap V ' \in \mathcal{P} \}$;\

  (b) $\overline{N3}_{j}^{\mathcal{P}}(V)$=  $(\underline{N3}_{j}^{\mathcal{P}}(V'))'$
  
   (c) $B^{\mathcal{P}}_{N3_{j}}(V)$= $\overline{N3}^{\mathcal{P}}_{j}(V) - \underline{N3}^{\mathcal{P}}_{j}(V) $\

  (d) $\sigma_{N3_j}^\mathcal{P}(V)= \frac{|\underline{N3}^{\mathcal{P}}_{j}(V)\cap V |}{|\overline{N3}^{\mathcal{P}}_{j}(V)\cup V |}$. \\

\end{D}

\begin{Pro} \label{Proposition 3.7} 
Let$\mathcal{P}$, $\mathcal{P}1$ be primals and $\Omega$be a binaryrelation on $\Xi$. Then, let V, W $\subseteq$  $\Xi$. Then, the subsequent is true :\

     $ (a)$ $V\subseteq W$ implies $\underline{N3}_{j}^{\mathcal{P}}(V)$$\subseteq$ $\underline{N3}_{j}^{\mathcal{P}}(W)$;\

    \indent \indent \indent \indent  \indent \indent $\overline{N3}_{j}^{\mathcal{P}}(V)$ $\subseteq $$\overline{N3}_{j}^{\mathcal{P}}(W)$;\

  $ (b)$ $\underline{N3}_{j}^{\mathcal{P}}(V \cup W)\supseteq$ $\underline{N3}_{j}^{\mathcal{P}}(V) \cup$ $\underline{N3}_{j}^{\mathcal{P}}(W)$;\

   \indent \indent   $\overline{N3}_{j}^{\mathcal{P}}(V \cup W)\supseteq$ $\overline{N3}_{j}^{\mathcal{P}}(V) \cup$ $\overline{N3}_{j}^{\mathcal{P}}(W)$;\

  $ (c)$ $\underline{N3}_{j}^{\mathcal{P}}(V \cap W)\subseteq$ $\underline{N3}_{j}^{\mathcal{P}}(V) \cap$ $\underline{N3}_{j}^{\mathcal{P}}(W)$;\

   \indent \indent   $\overline{N3}_{j}^{\mathcal{P}}(V \cap W)\subseteq$ $\overline{N3}_{j}^{\mathcal{P}}(V) \cap$ $\overline{N3}_{j}^{\mathcal{P}}(W)$;\

$(d)$ $\underline{N3}_{j}^{\mathcal{P}}(V)$ = $(\overline{N3}_{j}^{\mathcal{P}}(V '))'$

   \begin{proof} \

 $ (a)$ Let  $ \iota \in \underline{N3}_{j}^{\mathcal{P}}(V)$. Then there exists $\iota_1$ $\in$ $\Xi $ such that  $ \iota $ $\in $$w_j(\iota_1) \cap V ' \in \mathcal{P}$. Since  $V \subseteq W$ and  by the definition of primal, $ \iota$ $\in$ $w_j(\iota_1) \cap W ' \in \mathcal{P}$. So 
$\underline{N3}_{j}^{\mathcal{P}}(V)$ $\subseteq$ 
$\underline{N3}_{j}^{\mathcal{P}}(W)$. \

 $ (b)$ It originates directly from  $(a)$. \

 $ (c)$ It originates directly from  $(a)$. \

$(d)$ The proof is obvious.

\end{proof}
\end{Pro}

\begin{Pro} \label{Proposition 3.8} 
Let $\mathcal{P}$, $\mathcal{P}1$ be primals and $\Omega$ be a binary relation on $\Xi$. Then, let V$\subseteq$  $\Xi$. Then, the subsequent is true :\

$(a)$ If  $\mathcal{P}$ $\subseteq$ $\mathcal{P}1$ then 
$\underline{N3}_{j}^{\mathcal{P}}(V)$$\subseteq$ $\underline{N3}_{j}^{\mathcal{P}1}(V)$; \

\indent \indent  \indent \indent \indent  \indent
$\overline{N3}_{j}^{\mathcal{P}1}(V)$$\subseteq$ $\overline{N3}_{j}^{\mathcal{P}}(V)$ \

   $(b)$ $\underline{N3}_{j}^{\mathcal{P}\cup \mathcal{P}1}(V )$ = $\underline{N3}_{j}^{\mathcal{P}}(V) \cup$ $\underline{N3}_{j}^{\mathcal{P}1}(V)$;\

   \begin{proof} \

$(b)$ $\underline{N3}_{j}^{\mathcal{P}\cup \mathcal{P}1}(V ) $= $\bigcup_{\iota \in \Xi} \{ w_j(\iota) : w_j(\iota) \cap V ' \in \mathcal{P}\cup \mathcal{P}1 \}$ = $\bigcup_{\iota \in \Xi} \{ w_j(\iota) : w_j(\iota) \cap V ' \in \mathcal{P} \}$ $\cup$ $\bigcup_{\iota \in \Xi} \{ w_j(\iota) : w_j(\iota) \cap V ' \in \mathcal{P}1 \}$ = $\underline{N3}_{j}^{\mathcal{P}}(V ) $ $\cup$ $\underline{N3}_{j}^{\mathcal{P}1}(V ) $. \

\end{proof}
\end{Pro}

The equations in Proposition \ref{Proposition 3.7}(a), (b) and (c) might not hold true in general, as demonstrated by the example that follows. Furthermore, it suggests that Proposition \ref{Proposition 3.8} (a) might not deliver their contradictory meanings.

\begin{E}\label{Example 3.11} 
 Let $\Xi$ = $\{ \iota_1, \iota_2, \iota_3, \iota_4 \}$,  $\Omega$ = $\{ (\iota_1, \iota_1), (\iota_2, \iota_2),  (\iota_1, \iota_2),  (\iota_1, \iota_3)\}$, $\mathcal{P}$ = $\{\emptyset,  \{\iota_2 \}\}$ and $\mathcal{P}1$ = $\{\emptyset,  \{\iota_4 \}\}$.  \

$(a)$ Let $V$ = $\{  \iota_1, \iota_2  \}$ and $W$ = $\{  \iota_3\}$. Then 
$\underline{N3}_{a}^{\mathcal{P}}(V)$ = $\underline{N3}_{a}^{\mathcal{P}}(W)$ = $\{ \iota_2\}$ We gain $\underline{N3}_{a}^{\mathcal{P}}(V)$$\subseteq$ $\underline{N3}_{a}^{\mathcal{P}}(W)$ but it is not  $V \not \subseteq W$. \

$(b)$  Let $V$ = $\{  \iota_1, \iota_2  \}$ and $W$ = $\{  \iota_3\}$. Then
 $\underline{N3}_{a}^{\mathcal{P}}(V \cup W)$ = $\{  \iota_1, \iota_2 , \iota_3\}$. 
We gain  
$\underline{N3}_{a}^{\mathcal{P}}(V)$ $\cup$ $\underline{N3}_{a}^{\mathcal{P}}(W)$ $\neq$ 
$\underline{N3}_{a}^{\mathcal{P}}(V \cup W)$. \

$(c)$ Let $V$ = $\{  \iota_1, \iota_2  \}$ and $W$ = $\{  \iota_2, \iota_3\}$. Then 
$\overline{N3}_{a}^{\mathcal{P}}(V)$ = $\{ \iota_1, \iota_3, \iota_4 \}$, $ \overline{N3}_{j}^{\mathcal{P}}(W)$ = $\{ \iota_1, \iota_3, \iota_4 \}$  and $\overline{N3}_{a}^{\mathcal{P}}(V \cap W)$= $\emptyset$ .  We gain  $\overline{N3}_{a}^{\mathcal{P}}(V)$$\cap$ $\overline{N3}_{a}^{\mathcal{P}}(W)$$\neq$ $\overline{N3}_{a}^{\mathcal{P}}(V \cap W)$. \

$(d)$ Let $V$ = $\{  \iota_1, \iota_2, \iota_3  \}$ . Then 
$\underline{N3}_{a}^{\mathcal{P}}(V)$ = $\underline{N3}_{a}^{\mathcal{P}1}(V)$ = $\{  \iota_1, \iota_2, \iota_3  \}$  We gain $\underline{N3}_{a}^{\mathcal{P}}(V)$= $\underline{N3}_{a}^{\mathcal{P}1}(V)$ but it is not  $\mathcal{P} \not \subseteq \mathcal{P}1$. \

\end{E}

\begin{R} \label{Remark 3.3} 
 The current model lacks the following characteristics of the approximation operators in the Pawlak model. \

$(a)$ $\underline{N3}_{j}^{\mathcal{P}}(\Xi)$ = $\Xi$ and $\overline{N3}_{j}^{\mathcal{P}}(\emptyset)$ = $\emptyset$.

$(b)$ $\underline{N3}_{j}^{\mathcal{P}}(V) \subseteq V \subseteq \overline{N3}_{j}^{\mathcal{P}}(V)$; \

$(c)$  $\underline{N3}_{j}^{\mathcal{P}}(\underline{N3}_{j}^{\mathcal{P}}(V))= \underline{N3}_{j}^{\mathcal{P}}(V);$ \

 \indent \indent  $\overline{N3}_{j}^{\mathcal{P}}(\overline{N3}_{j}^{\mathcal{P}}(V))= \overline{N3}_{j}^{\mathcal{P}}(V);$  \

 $(d)$  $\underline{N3}_{j}^{\mathcal{P}}(V \cap W) =$ $\underline{N3}_{j}^{\mathcal{P}}(V) $
$\cap$ $\underline{N3}_{j}^{\mathcal{P}}(W)$;\

$ (e)$  $\overline{N3}_{j}^{\mathcal{P}}(V \cup W)=$ $\overline{N3}_{j}^{\mathcal{P}}(V)$
$ \cup$ $\overline{N3}_{j}^{\mathcal{P}}(W)$;\ 

$(f)$ $\overline{N3}_{j}^{\mathcal{P}}(\underline{N3}_{j}^{\mathcal{P}}(V))= \underline{N3}_{j}^{\mathcal{P}}(V)$;\

 \indent \indent  $\underline{N3}_{j}^{\mathcal{P}}(\overline{N3}_{j}^{\mathcal{P}}(V))= \overline{N3}_{j}^{\mathcal{P}}(V)$

\end{R}
\begin{R} \label{Remark 3.4} 
  The following properties are not always hold. \

$ (a)$  If $\mathcal{P}$ = $P(\Xi) - \{ \Xi \}$ then 

if  V $\neq$ $\Xi$ then $\underline{N3}_{j}^{\mathcal{P}}(V)$ = $\emptyset$ and 
if V $\subseteq$ $\Xi$ then $\overline{N3}_{j}^{\mathcal{P}}(V)$ = $\emptyset$  \

   $(b)$  If $ V^{ı}$ $\in$ $\mathcal{P}$  then $\underline{N3}_{j}^{\mathcal{P}}(V)$ = $\Xi$ \

  $(c)$ If V $\in$ $\mathcal{P}$  then $\overline{N3}_{j}^{\mathcal{P}}(V)$ =  $\emptyset$; 

\end{R}
\begin{E}\label{Example 3.12} 
Consider Example 3.11. \

$(a)$ $\underline{N3}_{a}^{\mathcal{P}}(\Xi)$ = $\{ \iota_1, \iota_2, \iota_3 \}$. \

$(b)$ Let $V '$ = $\{  \iota_1, \iota_3, \iota_4  \}$$\in$ $\mathcal{P}$. But $\underline{N3}_{a}^{\mathcal{P}}(V)$ =  $\{ \iota_1, \iota_2, \iota_3 \}$ $\neq$ $\Xi$. 

$(c)$  Let $V$ = $\{  \iota_1, \iota_3  \}$. Then $\underline{N3}_{a}^{\mathcal{P}}(V )$  = $\{ \iota_1, \iota_2, \iota_3 \}$. So   $\underline{N3}_{a}^{\mathcal{P}}(V) \not \subseteq V$.

\end{E}

\begin{E}\label{Example 3.13} 
 Let $\Xi$ = $\{ \iota_1, \iota_2, \iota_3\}$,  $\Omega$ = $\{ (\iota_1, \iota_1), (\iota_1, \iota_2)\}$ and $\mathcal{P}$ = $\{\emptyset,  \{\iota_1 \}, \{\iota_2 \}, \{\iota_3 \}, \{\iota_1, \iota_2 \}, 
\{\iota_1, \iota_3 \},  \{\iota_2, \iota_3 \} \}$.  \

$(a)$  $V$ = $\{  \iota_3 \}$ $\in$ $\mathcal{P}$  but  $\overline{N3}_{a}^{\mathcal{P}}(V)$ = $\{ \iota_3 \}$ $\neq$$\emptyset$.  \

$(b)$ Let $V$ = $\{  \iota_1, \iota_2 \}$. Then $\overline{N3}_{a}^{\mathcal{P}}(V))$ = $\{  \iota_3 \}$ and $\underline{N3}_{a}^{\mathcal{P}}(\overline{N3}_{a}^{\mathcal{P}}(V))$ == $\{  \iota_3 \}$. So $\underline{N3}_{a}^{\mathcal{P}}(\overline{N3}_{a}^{\mathcal{P}}(V))$
$\neq$$ \overline{N3}_{a}^{\mathcal{P}}(V)$.

\end{E}
\begin{Pro} \label{Proposition 3.9} 
The subsequent items are hold:\

$(a)$ $\underline{N3}_{u}^{\mathcal{P}}(V)$$\subseteq$ $\underline{N3}_{a}^{\mathcal{P}}(V)$
$\subseteq$ $\underline{N3}_{i}^{\mathcal{P}}(V)$;\\

 \indent \indent $\overline{N3}_{i}^{\mathcal{P}}(V)$$\subseteq$ $\overline{N3}_{a}^{\mathcal{P}}(V)$
$\subseteq$ $\overline{N3}_{u}^{\mathcal{P}}(V)$;\\

$(b)$ $\underline{N3}_{u}^{\mathcal{P}}(V)$$\subseteq$ $\underline{N3}_{b}^{\mathcal{P}}(V)$$\subseteq$ $\underline{N3}_{i}^{\mathcal{P}}(V)$;\\

 \indent \indent $\overline{N3}_{i}^{\mathcal{P}}(V)$$\subseteq$ $\overline{N3}_{b}^{\mathcal{P}}(V)$
$\subseteq$ $\overline{N3}_{u}^{\mathcal{P}}(V)$;\\

\begin{proof} \
It is obvious.
\end{proof}
\end{Pro}

\begin{C} \label{Corollary 3.7} 
The subsequent items are hold:\

$ (a)$ $B^{\mathcal{P}}_{N3_{i}}(V)$$\subseteq$$B^{\mathcal{P}}_{N3_{a}}(V)$$\subseteq$$B^{\mathcal{P}}_{N3_{u}}(V)$;\\

$ (b)$ $B^{\mathcal{P}}_{N3_{i}}(V)$$\subseteq$$B^{\mathcal{P}}_{N3_{b}}(V)$$\subseteq$$B^{\mathcal{P}}_{N3_{u}}(V)$.\\

\end{C}

\begin{C} \label{Corollary 3.8} 
Let$\mathcal{P}$ be primal and $\Omega$be abinary relation on $\Xi$. Then, let V
$\subseteq$  $\Xi$. Then, the subsequent is true :\\

(a) $\sigma_{N3_u}^\mathcal{P}(V) $$\leq$ $\sigma_{N3_a}^\mathcal{P}(V)$$\leq$ 
$\sigma_{N3_i}^\mathcal{P}(V)$ ;\\

(b) $\sigma_{N3_u}^\mathcal{P}(V) $$\leq$ $\sigma_{N3_b}^\mathcal{P}(V)$$\leq$ 
$\sigma_{N3_i}^\mathcal{P}(V)$.\\

\end{C}

\begin{C} \label{Corollary 3.9} 
    Let$\mathcal{P}$, $\mathcal{P}1$ be primals such that $\mathcal{P}$ $\supseteq$$\mathcal{P}1$  and $\Omega$ be a binary relation on $\Xi$. Then, let V $\subseteq$  $\Xi$. Then, the subsequent is true :\\

   $ (a)$ $B^{\mathcal{P}1}_{N3_{j}}(V)$$\subseteq$$B^{\mathcal{P}}_{N3_{j}}(V)$\

   (b) $\sigma_{N3_j}^{\mathcal{P}1}(V) $$\leq$ $\sigma_{N3_j}^\mathcal{P}(V)$.
\end{C}

\begin{E}\label{Example 3.14} 
Consider Example 3.13 and $\mathcal{P}1$ = $\{\emptyset,  \{\iota_1 \} \}$. 
 Let $V$ = $\{ \iota_1, \iota_3 \}$. Then $\sigma_{N3_b}^ \mathcal{P}(V)$ = $\frac{1}{2}$. but 
$\sigma_{N3_b}^{\mathcal{P}1}(V)$ = 0.
\end{E}

\subsection{PRIMAL-BASED GENERALIZED ROUGH SETS: THE FOURTH TECHNIQUE}
\begin{D}\label{Definition 3.4} 
Let $\mathcal{P}$  and $\Omega$ stand for the primal on a nonempty set $\Xi$ and the binary relation, respectively. The border region, accuracy, roughness, and enhanced operators (upper and lower) of a nonempty subset V of$\Xi$ generated from $\Omega$ and $\mathcal{P}$ are provided, respectively, by \

   (a)   $\overline{N4}_{j}^{\mathcal{P}}(V)=\bigcup_{\iota \in \Xi} \{ w_j(\iota) : w_j(\iota) \cap V  \not \in \mathcal{P}  \}$.\

  (b) $\underline{N4}_{j}^{\mathcal{P}}(V)$=  $(\overline{N4}_{j}^{\mathcal{P}}(V'))'$
  
   (c) $B^{\mathcal{P}}_{N4_{j}}(V)$= $\overline{N4}^{\mathcal{P}}_{j}(V) - \underline{N4}^{\mathcal{P}}_{j}(V) $\

  (d) $\sigma_{N4_j}^\mathcal{P}(V)= \frac{|\underline{N4}^{\mathcal{P}}_{j}(V)\cap V |}{|\overline{N4}^{\mathcal{P}}_{j}(V)\cup V |}$. \\

\end{D}

\begin{Pro} \label{Proposition 3.10} 
Let $\mathcal{P}$, $\mathcal{P}1$ be primals and $\Omega$ be a binary relation on $\Xi$. Then, let V, W $\subseteq$  $\Xi$. Then, the subsequent is true :\

$(a)$ $\underline{N4}_{j}^{\mathcal{P}}(\Xi)$ = $\Xi$ and $\overline{N4}_{j}^{\mathcal{P}}(\emptyset)$ = $\emptyset$. \

     $ (b)$ $V\subseteq W$ implies $\underline{N4}_{j}^{\mathcal{P}}(V)$$\subseteq$ $\underline{N4}_{j}^{\mathcal{P}}(W)$;\

    \indent \indent \indent \indent  \indent \indent $\overline{N4}_{j}^{\mathcal{P}}(V)$ $\subseteq $$\overline{N4}_{j}^{\mathcal{P}}(W)$;\

  $ (c)$ $\underline{N4}_{j}^{\mathcal{P}}(V \cup W)\supseteq$ $\underline{N4}_{j}^{\mathcal{P}}(V) \cup$ $\underline{N4}_{j}^{\mathcal{P}}(W)$;\

   \indent \indent   $\overline{N4}_{j}^{\mathcal{P}}(V \cup W)\supseteq$ $\overline{N4}_{j}^{\mathcal{P}}(V) \cup$ $\overline{N4}_{j}^{\mathcal{P}}(W)$;\

  $ (d)$ $\underline{N4}_{j}^{\mathcal{P}}(V \cap W)\subseteq$ $\underline{N4}_{j}^{\mathcal{P}}(V) \cap$ $\underline{N4}_{j}^{\mathcal{P}}(W)$;\

   \indent \indent   $\overline{N4}_{j}^{\mathcal{P}}(V \cap W)\subseteq$ $\overline{N4}_{j}^{\mathcal{P}}(V) \cap$ $\overline{N4}_{j}^{\mathcal{P}}(W)$;\

$(e)$ $\overline{N4}_{j}^{\mathcal{P}}(V)$ = $(\underline{N4}_{j}^{\mathcal{P}}(V '))'$

   \begin{proof} \

 $ (b)$ Let  $ \iota \not \in \underline{N4}_{j}^{\mathcal{P}}(W)$. Then there is not exists $\iota_1$ $\in$ $\Xi $ such that  $ \iota $ $\in $$w_j(\iota_1) \cap W \in \mathcal{P}$. Since  $V \subseteq W$ and  by the definition of primal, $ \iota$ $ \in$ $w_j(\iota_1) \cap V  \in \mathcal{P}$. So 
$\underline{N4}_{j}^{\mathcal{P}}(V)$ $\subseteq$ 
$\underline{N4}_{j}^{\mathcal{P}}(W)$.  \

 $ (c)$ It originates directly from  $(b)$. \

 $ (d)$ It originates directly from  $(b)$. \

$(e)$ The proof is obvious.

\end{proof}
\end{Pro}

\begin{Pro} \label{Proposition 3.11} 
Let$\mathcal{P}$, $\mathcal{P}1$ be primals and $\Omega$be abinary relation on $\Xi$. Then, let V$\subseteq$  $\Xi$. Then, the subsequent is true :\

$ (a)$  If $\mathcal{P}$ = $P(\Xi) - \{ \Xi \}$ then 

if  V $\neq$ $\Xi$ then  $\overline{N4}_{j}^{\mathcal{P}}(V)$ = $\emptyset$  and \

if  V $\neq$ $\emptyset$ then   $\underline{N4}_{j}^{\mathcal{P}}(V)$ = $\Xi$.\

$(b)$ If  $\mathcal{P}$ $\subseteq$ $\mathcal{P}1$ then 
$\underline{N4}_{j}^{\mathcal{P}}(V)$$\subseteq$ $\underline{N4}_{j}^{\mathcal{P}1}(V)$; \

\indent \indent  \indent \indent \indent  \indent
$\overline{N4}_{j}^{\mathcal{P}1}(V)$$\subseteq$ $\overline{N4}_{j}^{\mathcal{P}}(V)$ \

   $(c)$ $\underline{N4}_{j}^{\mathcal{P}\cup \mathcal{P}1}(V )$ = $\underline{N4}_{j}^{\mathcal{P}}(V) \cup$ $\underline{N4}_{j}^{\mathcal{P}1}(V)$;\

   \begin{proof} \
The proof is done similarly to the proof of Proposition 3.8.
\end{proof}
\end{Pro}

The equations in Proposition \ref{Proposition 3.10}(a), (b) and (c) might not hold true in general, as demonstrated by the example that follows. Furthermore, it suggests that Proposition \ref{Proposition 3.11} (b) might not deliver their contradictory meanings.

\begin{E}\label{Example 3.15} 
Consider Example 3.11 and $\mathcal{P}1$ = $\{\emptyset,   \{\iota_1 \} \}$. 

$(a)$ Let $V$ = $\{  \iota_1, \iota_2, \iota_3  \}$ and $W$ = $\{  \iota_4\}$. Then 
$\overline{N4}_{a}^{\mathcal{P}}(V)$ =  $\{  \iota_1, \iota_2, \iota_3  \}$  and  $\overline{N4}_{a}^{\mathcal{P}}(W)$ = $\emptyset$. We gain $\underline{N3}_{a}^{\mathcal{P}}(W)$$\subseteq$ $\underline{N4}_{a}^{\mathcal{P}}(V)$ but it is not  $W \not \subseteq V$. \

$(b)$  Let $V$ = $\{  \iota_1, \iota_2  \}$ and $W$ = $\{  \iota_2, \iota_3\}$. Then $\overline{N4}_{a}^{\mathcal{P}}(V)$ = $\overline{N4}_{a}^{\mathcal{P}}(W)$ =$\{  \iota_1, \iota_2, \iota_3  \}$  
 $\overline{N4}_{a}^{\mathcal{P}}(V \cap W)$ = $\emptyset$. 
We gain  
$\overline{N4}_{a}^{\mathcal{P}}(V)$ $\cap$ $\overline{N4}_{a}^{\mathcal{P}}(W)$ $\neq$ 
$\underline{N4}_{a}^{\mathcal{P}}(V \cap W)$. \

$(c)$ Let $V$ = $\{  \iota_1, \iota_2  \}$ and $W$ = $\{  \iota_3, \iota_4\}$. Then 
$\overline{N4}_{a}^{\mathcal{P}}(V)$ = $ \overline{N4}_{j}^{\mathcal{P}}(W)$ = $\{\iota_4 \}$,   and $\overline{N4}_{a}^{\mathcal{P}}(V \cup W)$= $\{  \iota_1, \iota_2, \iota_3  \}$.  We gain  $\overline{N4}_{a}^{\mathcal{P}}(V)$$\cup$ $\overline{N4}_{a}^{\mathcal{P}}(W)$$\neq$ $\overline{N4}_{a}^{\mathcal{P}}(V \cup W)$. \

$(d)$ Let $V$ = $\{  \iota_1, \iota_2  \}$. Then 
$\overline{N4}_{a}^{\mathcal{P}}(V)$ = $\overline{N4}_{a}^{\mathcal{P}1}(V)$ = $\{  \iota_1, \iota_2, \iota_3  \}$.  We gain $\underline{N4}_{a}^{\mathcal{P}}(V)$= $\underline{N4}_{a}^{\mathcal{P}1}(V)$ but it is not  $\mathcal{P} \not \subseteq \mathcal{P}1$. \

\end{E}

\begin{R} \label{Remark 3.5} 
  The current model lacks the following characteristics of the approximation operators in the Pawlak model. \

$(a)$ $\underline{N4}_{j}^{\mathcal{P}}(V) \subseteq V \subseteq \overline{N4}_{j}^{\mathcal{P}}(V)$; \

$(b)$  $\underline{N4}_{j}^{\mathcal{P}}(\underline{N4}_{j}^{\mathcal{P}}(V))= \underline{N4}_{j}^{\mathcal{P}}(V);$ \

 \indent \indent  $\overline{N4}_{j}^{\mathcal{P}}(\overline{N4}_{j}^{\mathcal{P}}(V))= \overline{N4}_{j}^{\mathcal{P}}(V);$  \

 $(c)$  $\underline{N4}_{j}^{\mathcal{P}}(V \cap W) =$ $\underline{N4}_{j}^{\mathcal{P}}(V) $
$\cap$ $\underline{N4}_{j}^{\mathcal{P}}(W)$;\

$ (d)$  $\overline{N4}_{j}^{\mathcal{P}}(V \cup W)=$ $\overline{N4}_{j}^{\mathcal{P}}(V)$
$ \cup$ $\overline{N4}_{j}^{\mathcal{P}}(W)$;\ 

$(e)$ $\overline{N4}_{j}^{\mathcal{P}}(\underline{N4}_{j}^{\mathcal{P}}(V))= \underline{N4}_{j}^{\mathcal{P}}(V)$;\

 \indent \indent  $\underline{N4}_{j}^{\mathcal{P}}(\overline{N4}_{j}^{\mathcal{P}}(V))= \overline{N4}_{j}^{\mathcal{P}}(V)$

\end{R}
\begin{R} \label{Remark 3.6} 
  The following properties are not always hold: \

   $(a)$  If $ V^{ı}$ $\in$ $\mathcal{P}$  then $\underline{N4}_{j}^{\mathcal{P}}(V)$ = $\Xi$ \

  $(b)$ If V $\in$ $\mathcal{P}$  then $\overline{N4}_{j}^{\mathcal{P}}(V)$ =  $\emptyset$; 

\end{R}
\begin{E}\label{Example 3.16} 
Consider Example 3.11. \

$(a)$  Let $V$ = $\{  \iota_2, \iota_4  \}$. Then $\overline{N4}_{a}^{\mathcal{P}}(V)$ = $\emptyset$ but  V $ \not \in$ $\mathcal{P}$\

$(b)$  Let $V$ = $\{  \iota_1, \iota_3  \}$. Then $\underline{N4}_{a}^{\mathcal{P}}(V )$  = $\Xi$. So   $\underline{N4}_{a}^{\mathcal{P}}(V) \not \subseteq V$. \

$(c)$  Let $V$ = $\{  \iota_1, \iota_3  \}$. Then $\underline{N4}_{a}^{\mathcal{P}}(V )$  = $\Xi$ and  $\overline{N4}_{j}^{\mathcal{P}}(\underline{N4}_{j}^{\mathcal{P}}(V))$ = $\{  \iota_1, \iota_2, \iota_3  \}$.  So   $\overline{N4}_{j}^{\mathcal{P}}(\underline{N4}_{j}^{\mathcal{P}}(V)) \neq \underline{N4}_{j}^{\mathcal{P}}(V)$;\ \

\end{E}

\begin{E}\label{Example 3.17} 
Consider Example 3.4.   Let $V$ = $\{  \iota_2, \iota_3 \}$. Then $\overline{N4}_{a}^{\mathcal{P}}(V))$ =
 $\{  \iota_1, \iota_2, \iota_3 \}$ and $\overline{N4}_{a}^{\mathcal{P}}(\overline{N4}_{a}^{\mathcal{P}}(V))$ = $\Xi$. So $\overline{N4}_{a}^{\mathcal{P}}(\overline{N4}_{a}^{\mathcal{P}}(V))$
$\neq$$ \overline{N4}_{a}^{\mathcal{P}}(V)$.

\end{E}
\begin{Pro} \label{Proposition 3.12} 
The subsequent items are hold:\

$(a)$ $\underline{N4}_{u}^{\mathcal{P}}(V)$$\subseteq$ $\underline{N4}_{a}^{\mathcal{P}}(V)$
$\subseteq$ $\underline{N4}_{i}^{\mathcal{P}}(V)$;\\

 \indent \indent $\overline{N4}_{i}^{\mathcal{P}}(V)$$\subseteq$ $\overline{N4}_{a}^{\mathcal{P}}(V)$
$\subseteq$ $\overline{N4}_{u}^{\mathcal{P}}(V)$;\\

$(b)$ $\underline{N4}_{u}^{\mathcal{P}}(V)$$\subseteq$ $\underline{N4}_{b}^{\mathcal{P}}(V)$$\subseteq$ $\underline{N4}_{i}^{\mathcal{P}}(V)$;\\

 \indent \indent $\overline{N4}_{i}^{\mathcal{P}}(V)$$\subseteq$ $\overline{N4}_{b}^{\mathcal{P}}(V)$
$\subseteq$ $\overline{N4}_{u}^{\mathcal{P}}(V)$;\\

\begin{proof} \
It is obvious.
\end{proof}
\end{Pro}

\begin{C} \label{Corollary 3.10} 
The subsequent items are hold:\

$ (a)$ $B^{\mathcal{P}}_{N4_{i}}(V)$$\subseteq$$B^{\mathcal{P}}_{N4_{a}}(V)$$\subseteq$$B^{\mathcal{P}}_{N4_{u}}(V)$;\\

$ (b)$ $B^{\mathcal{P}}_{N4_{i}}(V)$$\subseteq$$B^{\mathcal{P}}_{N4_{b}}(V)$$\subseteq$$B^{\mathcal{P}}_{N4_{u}}(V)$.\\

\end{C}

\begin{C} \label{Corollary 3.11} 
Let $\mathcal{P}$ be primal and $\Omega$ be a binary relation on $\Xi$. Then, let V
$\subseteq$  $\Xi$. Then, the subsequent is true :\\

(a) $\sigma_{N4_u}^\mathcal{P}(V) $$\leq$ $\sigma_{N4_a}^\mathcal{P}(V)$$\leq$ 
$\sigma_{N4_i}^\mathcal{P}(V)$ ;\\

(b) $\sigma_{N4_u}^\mathcal{P}(V) $$\leq$ $\sigma_{N4_b}^\mathcal{P}(V)$$\leq$ 
$\sigma_{N4_i}^\mathcal{P}(V)$.\\

\end{C}

\begin{C} \label{Corollary 3.12} 
    Let$\mathcal{P}$, $\mathcal{P}1$ be primals such that $\mathcal{P}$ $\supseteq$$\mathcal{P}1$  and $\Omega$ be a binary relation on $\Xi$. Then, let V $\subseteq$  $\Xi$. Then, the subsequent is true :\\

   $ (a)$ $B^{\mathcal{P}1}_{N4_{j}}(V)$$\subseteq$$B^{\mathcal{P}}_{N4_{j}}(V)$;\

    $ (b)$ $\sigma_{N4_j}^{\mathcal{P}1}(V) $$\leq$ $\sigma_{N4_j}^\mathcal{P}(V)$.
\end{C}

\begin{E}\label{Example 3.18} 
Consider Example 3.1. $\mathcal{P}1$ = $\{\emptyset,  \{\iota_1 \} \}$. 
 Let $V$ = $\{ \iota_1, \iota_2 \}$. Then $\sigma_{N4_a}^ \mathcal{P}(V)$ =1 and
$\sigma_{N4_a}^{\mathcal{P}1}(V)$ =$\emptyset$.
\end{E}
\section{EXAMINE AND CONTRAST THE ADVANTAGES OF THE PROPOSED METHODS WITH THOSE OF THE PREVIOUS ONES}
\subsection{Extend And Compare The Suggested Methods With Relatıve Accuracy Measurements Of Subsets}
\begin{T} \label{Theorem 4.1} 
Let$\mathcal{P}$  and $\Omega$  stand for the primal on a nonempty set $\Xi$ and the binary relation, respectively. Let $V \subseteq \Xi$  \

$(a)$  $\underline{N2}_{j}^{\mathcal{P}}(V)$$\subseteq$ $\underline{N1}_{j}^{\mathcal{P}}(V)$; \

$(b)$  $\overline{N1}_{j}^{\mathcal{P}}(V)$$\subseteq$ $\overline{N2}_{j}^{\mathcal{P}}(V)$; \

$ (c)$  $B^{\mathcal{P}}_{N1_{j}}(V)$$\subseteq$$B^{\mathcal{P}}_{N2_{j}}(V)$;\

$(d)$  $\sigma_{N1_j}^{\mathcal{P}}(V) $= $\sigma_{N2_j}^\mathcal{P}(V)$.
\begin{proof}
Definitions 3.1 and 3.2 directly follow to provide the proof.
\end{proof}
\end{T}

The previous theorem's inclusion and less than relations are suitable. \

\begin{E}\label{Example 4.1} 
 Consider Example 3.1.  Let $V$ = $\{  \iota_2, \iota_3,  \iota_4 \}$. Then $\underline{N1}_{a}^{\mathcal{P}}(V)$ = $\emptyset$ and $\underline{N2}_{a}^{\mathcal{P}}(V))$ = $\{  \iota_2, \iota_3,  \iota_4 \}$. 
\end{E}
\begin{E}\label{Example 4.2} 
 Consider Example 3.6.  Let $V$ = $\{  \iota_1, \iota_3 \}$. Then  $B^{\mathcal{P}}_{N1_{a}}(V)$ = $\emptyset$  and $B^{\mathcal{P}}_{N2_{a}}(V)$ = $\{ \iota_3\}$. 
\end{E}
\begin{T} \label{Theorem 4.2} 
Let$\mathcal{P}$  and $\Omega$  stand for the primal on a nonempty set $\Xi$ and the reflexive  relation, respectively. Let $V \subseteq \Xi$  \

$(a)$  $\underline{N2}_{j}^{\mathcal{P}}(V)$$\subseteq$ $\underline{N1}_{j}^{\mathcal{P}}(V)$$\subseteq$  $\underline{N3}_{j}^{\mathcal{P}}(V)$; \

$(b)$ $\overline{N3}_{j}^{\mathcal{P}}(V)$$\subseteq$ $\overline{N1}_{j}^{\mathcal{P}}(V)$$\subseteq$ $\overline{N2}_{j}^{\mathcal{P}}(V)$; \

$ (c)$ $B^{\mathcal{P}}_{N3_{j}}(V)$ $\subseteq$ $B^{\mathcal{P}}_{N1_{j}}(V)$$\subseteq$$B^{\mathcal{P}}_{N2_{j}}(V)$;\

$(d)$  $\sigma_{N2_j}^{\mathcal{P}}(V) $$\leq$ $\sigma_{N1_j}^\mathcal{P}(V)$ $\leq$ $\sigma_{N3_j}^\mathcal{P}(V)$.
\begin{proof}
$(a)$ Let $\iota$ $\in$ $\underline{N1}_{j}^{\mathcal{P}}(V)$. Then  $w_j(\iota) \cap V'   \in \mathcal{P}$. Hence $w_j(\iota)$ $\in$ $\underline{N3}_{j}^{\mathcal{P}}(V)$. Since  $\Omega$  is  reflexive  relation,   $\iota$ $\in$ $\underline{N3}_{j}^{\mathcal{P}}(V)$. So $\underline{N1}_{j}^{\mathcal{P}}(V)$$\subseteq$  $\underline{N3}_{j}^{\mathcal{P}}(V)$. \

$(b)$ By Theorem 4.1, we have  $\overline{N1}_{j}^{\mathcal{P}}(V)$ $\subseteq$ 
$\overline{N2}_{j}^{\mathcal{P}}(V)$.   Let $\iota$ $\in$ $\overline{N3}_{j}^{\mathcal{P}}(V)$.  Then   $\iota$ $\not \in$  $(\underline{N3}_{j}^{\mathcal{P}}(V'))$ by Definition 3.3.  So   $w_j(\iota) \cap (V')'  \not \in \mathcal{P}$. We obtain  $\iota$ $\in$ $\overline{N1}_{j}^{\mathcal{P}}(V)$. \

$(c)$-$(d)$ Simple explanations from $(a)$ and $(b)$.
\end{proof}
\end{T}

The previous theorem's inclusion and less than relations are suitable. \

\begin{E}\label{Example 4.3} 
Consider Example 3.4.  Let $V$ = $\{  \iota_1, \iota_3 \}$.  Then $\underline{N3}_{a}^{\mathcal{P}}(V)$ =  $\{  \iota_1, \iota_2, \iota_3\}$ and $\underline{N1}_{a}^{\mathcal{P}}(V)$ =  $\{  \iota_1, \iota_2 \} $.  Besides $\overline{N3}_{a}^{\mathcal{P}}(V)$ =  $\{  \iota_1\}$ and $\overline{N1}_{a}^{\mathcal{P}}(V)$ =  $\{  \iota_1, \iota_4 \} $. Then we obtain; \

$(a)$  $\underline{N3}_{a}^{\mathcal{P}}(V)$$ \not \subseteq$  $\underline{N1}_{a}^{\mathcal{P}}(V)$; \

$(b)$ $\overline{N1}_{a}^{\mathcal{P}}(V)$$\not \subseteq$ $\overline{N3}_{a}^{\mathcal{P}}(V)$; \

$(c)$  $\sigma_{N3_a}^\mathcal{P}(V)$ =1 $\not \leq$ $\frac{1}{3}$=$\sigma_{N1_a}^\mathcal{P}(V)$.
\end{E}
\begin{T} \label{Theorem 4.2} 
Let$\mathcal{P}$  and $\Omega$  stand for the primal on a nonempty set $\Xi$ and the reflexive  relation, respectively. Let $V \subseteq \Xi$.  \

$(a)$  $\underline{N4}_{j}^{\mathcal{P}}(V)$$\subseteq$ $\underline{N1}_{j}^{\mathcal{P}}(V)$$\subseteq$  $\underline{N3}_{j}^{\mathcal{P}}(V)$; \

$(b)$ $\overline{N3}_{j}^{\mathcal{P}}(V)$$\subseteq$ $\overline{N1}_{j}^{\mathcal{P}}(V)$$\subseteq$ $\overline{N4}_{j}^{\mathcal{P}}(V)$; \

$ (c)$ $B^{\mathcal{P}}_{N3_{j}}(V)$ $\subseteq$ $B^{\mathcal{P}}_{N1_{j}}(V)$$\subseteq$$B^{\mathcal{P}}_{N4_{j}}(V)$;\

$(d)$  $\sigma_{N4_j}^{\mathcal{P}}(V) $$\leq$ $\sigma_{N1_j}^\mathcal{P}(V)$ $\leq$ $\sigma_{N3_j}^\mathcal{P}(V)$.
\begin{proof}
It is done similarly to Theorem 4.2.
\end{proof}
\end{T}
\begin{E}\label{Example 4.4} 
Consider Example 3.4.  Let $V$ = $\{  \iota_1,  \iota_4 \}$. Then $\overline{N4}_{a}^{\mathcal{P}}(V)$ =  $\Xi$ and $\overline{N1}_{a}^{\mathcal{P}}(V)$ =  $\{   \iota_1, \iota_3, \iota_4 \} $.  Besides $\underline{N4}_{a}^{\mathcal{P}}(V)$ =  $\{  \iota_4\}$ and $\underline{N1}_{a}^{\mathcal{P}}(V)$ =  $\{  \iota_3, \iota_4 \} $. Then we obtain; \

$(a)$  $\underline{N1}_{a}^{\mathcal{P}}(V)$$ \not \subseteq$  $\underline{N4}_{a}^{\mathcal{P}}(V)$; \

$(b)$ $\overline{N4}_{a}^{\mathcal{P}}(V)$$\not \subseteq$ $\overline{N1}_{a}^{\mathcal{P}}(V)$; \

$(c)$  $\sigma_{N1_a}^\mathcal{P}(V)$ =$\frac{1}{3}$ $\not \leq$ $\frac{1}{4}$=$\sigma_{N4_a}^\mathcal{P}(V)$.
\end{E}
Below is an illustration of why the fourth and third techniques cannot be compared to one another. \\

\begin{E}\label{Example 4.5} 
Consider Example 3.4.  Let $V$ = $\{  \iota_1,  \iota_4 \}$ and $W$ = $\{  \iota_2 \}$. . Then $\underline{N1}_{a}^{\mathcal{P}}(V)$ =  $\{  \iota_3, \iota_4 \} $ and $\underline{N}_{a}(V)$ =  $\{ \iota_4 \} $.  Besides 
$\overline{N1}_{a}^{\mathcal{P}}(W)$ =  $\emptyset$ and $\overline{N}_{a}(W)$ =  $\{  \iota_1, \iota_2 \} $. Afterward, we acquire; \

$(a)$  $\underline{N1}_{a}^{\mathcal{P}}(V)$$ \not \subseteq$  $\underline{N}_{a}(V)$; \

$(b)$ $\overline{N}_{a}(W)$$\not \subseteq$ $\overline{N1}_{a}^{\mathcal{P}}(W)$; \

$(c)$  $\sigma_{N1_a}^\mathcal{P}(W)$ =$1$ $\not \leq$ $0$ = $\sigma_{N_a}(W)$.
\end{E}
\begin{T} \label{Theorem 4.4} 
Let$\mathcal{P}$  and $\Omega$  stand for the primal on a nonempty set $\Xi$ and the binary relation, respectively. Let $V \subseteq \Xi$  \

$(a)$  $\underline{N}_{j}(V)$$\subseteq$ $\underline{N1}_{j}^{\mathcal{P}}(V)$; \

$(b)$  $\overline{N1}_{j}^{\mathcal{P}}(V)$$\subseteq$ $\overline{N}_{j}(V)$; \

$(c)$  $\sigma_{N_j}^{\mathcal{P}}(V) $$\leq$ $\sigma_{N1_j}^\mathcal{P}(V)$.
\begin{proof}
Definitions 3.1 and 2.4 directly follow to provide the proof.
\end{proof}
\end{T}
The above theorem's converse isn't always valid. \\

\begin{R} \label{remark 4.1}
A comparison of these four different techniques that we have defined in Table 1 is given, using to Example \ref{Example 3.4} by reflexive relation. 
 \end{R}

\begin{sidewaystable}[http]
\caption{A comparison of these four different techniques}\label{table1}.
\medskip
\centering\renewcommand{\arraystretch}{2}
\tiny
\begin{tabular}{|c||c|c|c||c|c|c||c|c|c||c|c|c|} \hline  \hline  \hline  \hline  \hline  \hline  \hline  \hline  \hline  \hline  \hline  \hline  \hline  \hline  \hline  \hline
V &\multicolumn{3}{c||}{$the first technique$}&\multicolumn{3}{c||}{$the second technique$}&\multicolumn{3}{c||}{$the third technique$}&\multicolumn{3}{c|}{$the fourth technique$} \\ \cline{2-13}
&$\underline{N1}_{a}^{\mathcal{P}}(V)$&$\overline{N1}_{a}^{\mathcal{P}}(V)$ & $
\sigma_{N1_a}^\mathcal{P}(V)$ &$\underline{N2}_{a}^{\mathcal{P}}(V)$&$\overline{N2}_{a}^{\mathcal{P}}(V)$ & $
\sigma_{N2_a}^\mathcal{P}(V)$ &$\underline{N3}_{a}^{\mathcal{P}}(V)$&$\overline{N3}_{a}^{\mathcal{P}}(V)$ & $
\sigma_{N3_a}^\mathcal{P}(V)$
&$\underline{N4}_{a}^{\mathcal{P}}(V)$&$\overline{N4}_{a}^{\mathcal{P}}(V)$ & $
\sigma_{N4_a}^\mathcal{P}(V)$ \\
\hline\hline
$\{\iota_1\}$ &$\emptyset$&$\{\iota_1, \iota_3\}$&$0$&$\emptyset$&$\{\iota_1, \iota_3\}$&$0$&$\emptyset$&$\{\iota_1\}$&$0$&
$\emptyset$&$\Xi$&$0$ \\
\hline
$\{\iota_2\}$ &$\{\iota_2, \iota_3\}$&$\emptyset$&$1$&$\{\iota_2, \iota_3\}$&$\{\iota_2\}$&$1$&$\{\iota_2, \iota_3\}$&$\emptyset$&$1$&
$\emptyset$&$\emptyset$&$0$ \\
\hline
$\{\iota_3\}$ &$\{\iota_2\}$&$\emptyset$&$0$&$\{\iota_2 \}$&$\{\iota_3\}$&$0$&$\{\iota_2, \iota_3\}$&$\emptyset$&$1$&
$\emptyset$&$\emptyset$&$0$ \\
\hline
$\{\iota_4\}$ &$\{\iota_3\}$&$\{\iota_3, \iota_4\}$&$0$&$\{\iota_3 \}$&$\{\iota_3, \iota_4\}$&$0$&$\{\iota_3, \iota_4\}$&$\{\iota_4\}$&$1$&
$\emptyset$&$\{\iota_1, \iota_3, \iota_4\}$&$0$ \\
\hline
$\{\iota_1, \iota_2\}$ &$\{\iota_1, \iota_2\}$&$\{\iota_1, \iota_4\}$&$ \frac{2}{3}$&$\{\iota_1, \iota_2\}$&$\{\iota_1, \iota_2, \iota_4\}$&$\frac{2}{3}$&$\{\iota_1, \iota_2, \iota_3\}$&$\{\iota_1\}$&$1$&
$\{\iota_2\}$&$\Xi$&$\frac{1}{2}$ \\
\hline
$\{\iota_1, \iota_3\}$ &$\{\iota_1, \iota_2\}$&$\{\iota_1, \iota_4\}$&$\frac{1}{3}$&$\{\iota_1, \iota_2\}$&$\{\iota_1, \iota_3, \iota_4\}$&$\frac{1}{3}$&$\{\iota_1, \iota_2, \iota_3\}$&$\{\iota_1\}$&$1$&
$\{\iota_2\}$&$\Xi$&$0$ \\
\hline
$\{\iota_1, \iota_4\}$ &$\{\iota_3, \iota_4\}$&$\{\iota_1, \iota_3,  \iota_4\}$&$\frac{1}{3}$&$\{\iota_3, \iota_4\}$&$\{\iota_1, \iota_3, \iota_4\}$&$\frac{1}{3}$&$\{\iota_1, \iota_3, \iota_4\}$&$\{\iota_1,  \iota_4\}$&$1$&
$\{\iota_4\}$&$\Xi$&$\frac{1}{4}$ \\
\hline
$\{\iota_2, \iota_3\}$ &$\{\iota_2\}$&$\{\iota_1, \iota_2 \}$&$\frac{1}{3}$&$\{\iota_2\}$&$\{\iota_1, \iota_2, \iota_3\}$&$\frac{1}{3}$&$\{\iota_2, \iota_3 \}$&$\{\iota_2 \}$&$1$&
$\emptyset$&$\{\iota_1, \iota_2,  \iota_3\}$&$0$ \\
\hline
$\{\iota_2, \iota_4\}$ &$\{\iota_2, \iota_3\}$&$\{\iota_3, \iota_4 \}$&$\frac{1}{3}$&$\{\iota_2, \iota_3\}$&$\{\iota_2, \iota_3, \iota_4\}$&$\frac{1}{3}$&$\{\iota_2, \iota_3, \iota_4 \}$&$\{\iota_4 \}$&$1$&
$\emptyset$&$\{\iota_1, \iota_3,  \iota_4\}$&$0$ \\
\hline
$\{\iota_3, \iota_4\}$ &$\{\iota_2, \iota_3\}$&$\{\iota_3, \iota_4 \}$&$\frac{1}{2}$&$\{\iota_2, \iota_3\}$&$\{\iota_3, \iota_4\}$&$\frac{1}{2}$&$\{\iota_2, \iota_3, \iota_4 \}$&$\{\iota_4 \}$&$1$&
$\emptyset$&$\{\iota_1, \iota_3,  \iota_4\}$&$0$ \\
\hline
$\{\iota_1, \iota_2, \iota_3\}$ &$\{\iota_1, \iota_2\}$&$\{\iota_1, \iota_2, \iota_4 \}$&$\frac{1}{2}$&$\{\iota_1, \iota_2\}$&$\Xi$&$\frac{1}{2}$&$\{\iota_1, \iota_2, \iota_3 \}$&$\{\iota_1, \iota_2 \}$&$1$&
$\{\iota_2\}$&$\Xi$&$\frac{1}{3}$ \\
\hline
$\{\iota_1, \iota_2, \iota_4\}$ &$\Xi$&$\{\iota_1, \iota_3, \iota_4 \}$&$\frac{3}{4}$&$\Xi$&$\Xi$&$\frac{3}{4}$&$\Xi$&$\{\iota_1, \iota_4 \}$&$1$&
$\Xi$&$\Xi$&$\frac{3}{4}$\\
\hline
$\{\iota_1,\iota_3,\iota_4\}$&$\Xi$&$\{\iota_1, \iota_4 \}$&$1$&$\Xi$&$\{\iota_1, \iota_3, \iota_4 \}$&$1$&$\Xi$&$\{\iota_1,\iota_4\}$&$1$&
$\Xi$&$\Xi$&$\frac{3}{4}$ \\
\hline
$\{\iota_2,\iota_3,\iota_4\}$&$\{\iota_2, \iota_4 \}$&$\Xi$&$\frac{1}{2}$&$\{\iota_2, \iota_4 \}$&$\Xi$&$\frac{1}{2}$&$\{\iota_2,\iota_3,\iota_4\}$&$\Xi$&$1$&
$\emptyset$&$\Xi$&$0$ \\
\hline
$\Xi$&$\Xi$&$\Xi$&$1$&$\Xi$&$\Xi$&$1$&
$\Xi$&$\Xi$&$1$&
$\Xi$&$\Xi$&$1$ \\
\hline
\end{tabular}
\end{sidewaystable}
\section{Applications}
\label{Sec:5}
Using the knowledge framework described above, we apply novel techniques presented in Section 3 to reduce the characteristics and identify significant indicators of breast cancer in females \cite{Sayed} in this segment.
Let $\Xi$ be a collection of goals, and let $\{1, 2, 3, 4, 5, 6 \}$ represent the six patients on the list. The characteristics are V = $\{A1, A2, A3, A4, A5 \}$, which stand for breast lump, inverted nipple, rashes, discharge from the nipple, and armpit swelling, in that order. Table 2 describes how individuals should consult a physician when they experience one or more breast cancer symptoms.
\begin{table}[h!]
\caption{Multi-Valued Information System (MVIS)} \label{t2}
\medskip
\centering\renewcommand{\arraystretch}{1.2}
\small
\begin{tabular}{c c c c c c c c } \hline
Person  & A1 &  A2    & A3 &  A4 & A5 &   Decision  \\  \hline
1       & Yes   &   Yes   & Yes &Yes &No    & Yes    \\
2   & Yes   &   Yes   & Yes &Yes &No    & Yes     \\
3     & No   &   Yes   & No &Yes &Yes   & No    \\
4     & Yes   &   No   & Yes &No &Yes    & No    \\
5     & No   &   Yes   & No &Yes &Yes    & No      \\
6    & Yes   &   Yes   & No &Yes &Yes    & Yes    \\
 \hline
\end{tabular}
\end{table}
Based on Table 1, we can infer each patient's symptoms as follows: \\
$V(1)$ =  $V(2)$= $\{A1, A2, A3, A4 \}$, $V(3)$ = $V(5)$= $\{ A2, A4, A5\}$, $V(4)$ =  $\{ A1, A3, A5\}$, and $V(6)$ =  $\{ A1, A2, A4, A5\}$. \\
Thus, the relation that represents this system can be given by: $m$$\Omega$$n$ $\Leftrightarrow$ V(m) $\subseteq$ V(n) , for each  x, y $\in \Xi$. Then $\Omega$ = $\{(1, 1), (2,2), (3, 3), (4,4), (5,5), (6, 6),(1,2), \\ (2, 1), (3,5), (3,6), (5, 3), (5, 6)\}$.\

Finally, we consider the primal is $\mathcal{P}=\{\emptyset,  \{3\}, \{5\}, \{6\}, \{3, 5\} \}$.  \\
Consequently, our techniques can be considered powerful tools for making decisions in the rough set theory and can benefit from the identification and extraction of concealed details in data collected through practical applications. Our techniques are the most effective means of evaluating the accuracy and ambiguities of the sets. As an illustration, we look at the subsets  K =$\{1, 2, 6\}$ and L = $\{3, 4, 5\}$, which stand for the set of infected brain cancer and uninfected brain cancer, respectively. Next, the approximations of them obtained from the current approaches in this study (Definition 3.3 ) and the Allam's methods in (Definition 2.3 ) are provided as follows, respectively: \\
Allam's method \cite{Allam1, Allam2}
The approximations for infected set K are: \\
$\underline{N}_{a}(K)$= $\{1, 2, 6\}$ \\
$\overline{N}_{a}(K)$= $\{1, 2, 3, 5, 6\}$ \\
 Thus $\sigma_{N_a}(K)$ =$\frac{3}{5}$. \\
Accordingly K is rough (not definable) set  defined by Allam's  method. \

In a comparable way, the estimates for the uninfected set L are: \\
$\underline{N}_{a}(L)$= $\{4\}$ \\
$\overline{N}_{a}(L)$= $\{3, 4, 5, 6\}$ \\
 Thus $\sigma_{N_a}(L)$ =$\frac{1}{4}$. \\
Accordingly L is rough (not definable) set. \ 

Our's methods
The approximations for infected set K are: \\
 $\underline{N3}_{a}^{\mathcal{P}}(K)$=$\{1, 2, 3, 5, 6\}$  \\
 $\overline{N3}_{a}^{\mathcal{P}}(K)$=$\{1, 2\}$  \\
 Thus $\sigma_{N3_a}^{\mathcal{P}}(K) $ =1. \\
As a result, K is precisely defined by our methods.  \

In a comparable way, the estimates for noninfected set in set L are: \\
The approximations for infected set L are: \\
 $\underline{N3}_{a}^{\mathcal{P}}(L)$=$\{3, 4\}$  \\
 $\overline{N3}_{a}^{\mathcal{P}}(L)$=$\{3, 4, 5 \}$  \\
 Thus  $\sigma_{N3_a}^{\mathcal{P}}(L) $ =1. \\
As a result, L is precisely defined by our methods. \\

\end{document}